\documentclass[journal]{IEEEtran}
\usepackage{etoolbox}
\makeatletter
\patchcmd{\@makecaption}
  {\scshape}
  {}
  {}
  {}
\makeatother

\usepackage[breaklinks,colorlinks,
  citecolor=blue,
  urlcolor=blue]{hyperref}
\usepackage{enumitem}
\usepackage{cuted}
\usepackage{cite}
\ifCLASSINFOpdf
   \usepackage[pdftex]{graphicx}
\else
  \usepackage{graphicx}
\fi
\graphicspath{{figures/}}

\usepackage{amsmath}
\usepackage{empheq}
\usepackage{bbm}
\usepackage{amssymb}
\usepackage{mathrsfs}  
\usepackage{amsthm}
\usepackage{bm}

\usepackage{array}

\usepackage{url}

\theoremstyle{plain}
\newtheorem{theorem}{Theorem}[section]
\newtheorem{proposition}[theorem]{Proposition}

\theoremstyle{definition}

\theoremstyle{remark}
\newtheorem{remark}[theorem]{Remark}

\newcommand{\bfi}{\bfseries\itshape}
\newcommand{\vertiii}[1]{{\left\vert\kern-0.25ex\left\vert\kern-0.25ex\left\vert #1 
    \right\vert\kern-0.25ex\right\vert\kern-0.25ex\right\vert}}
\newsavebox{\savepar}

\DeclareMathOperator*{\Motimes}{\text{\raisebox{0.25ex}{\scalebox{0.8}{$\bigotimes$}}}}

\begin{document}

\title{Reservoir kernels and Volterra series}

\author{Lukas~Gonon, Lyudmila~Grigoryeva, and
        Juan-Pablo~Ortega
\thanks{L. Gonon is affiliated with the University of St. Gallen, Switzerland, and is also an Honorary Senior Lecturer at Imperial College London, UK. L. Grigoryeva is with the University of St. Gallen, Switzerland, and is also an Honorary Associate Professor at the University of Warwick, UK. J.-P. Ortega is with the  Nanyang Technological University, Singapore.}
}


\markboth{Gonon, Grigoryeva, and Ortega: Reservoir kernels and Volterra series}%
{Gonon, Grigoryeva, and Ortega: Reservoir kernels and Volterra series}
%


\maketitle

\begin{abstract}
A universal kernel is constructed whose sections approximate any causal and time-invariant filter in the fading memory category with inputs and outputs in a finite-dimensional Euclidean space. This kernel is built using the reservoir functional associated with a state-space representation of the Volterra series expansion available for any analytic fading memory filter, and it is hence called the Volterra reservoir kernel. Even though the state-space representation and the corresponding reservoir feature map are defined on an infinite-dimensional tensor algebra space, the kernel map is characterized by explicit recursions that are readily computable for specific data sets when employed in estimation problems using the representer theorem. The empirical performance of the Volterra reservoir kernel is showcased and compared to other standard static and sequential kernels in a multidimensional and highly nonlinear learning task for the conditional covariances of financial asset returns.
\end{abstract}

\begin{IEEEkeywords}
Recurrent neural network, reservoir computing, fading memory property, recurrent kernel Hilbert space, RKHS, kernel methods, Volterra series expansion, polynomial regression, signature process.
\end{IEEEkeywords}

%
\IEEEpeerreviewmaketitle

\section{Introduction}
%
%
%
%
\IEEEPARstart{R}{eproducing} kernels and {\bfi  reproducing kernel Hilbert spaces (RKHS)}~\cite{aronszajn1950theory} have become one of the most important embedding methods in the development of machine learning algorithms  for which powerful numerical and theoretical tools have been developed~\cite{schoelkopf:smola:book, Hofmann2008}. Even though the spectrum of applications in which these techniques have been employed is very large, most research activity has been concentrated on static classification, regression and pattern recognition tasks. 

An important feature that arises when using reproducing kernels is the possibility of reducing the estimation of nonlinear data {dependencies} to linear regression problems via the so-called Representer Theorem~\cite{Mohri:learning:2012}. A similar goal has been pursued in a dynamical framework in a field collectively known as {\bfi  reservoir computing (RC)}~\cite{jaeger2001, maass1, Jaeger04, maass2}. RC aims at approximating nonlinear input/output systems using randomly generated state-space systems in which only a linear readout is estimated. It has been theoretically established that this is indeed possible in a variety of deterministic and stochastic contexts~\cite{RC6, RC7, RC8, RC20, RC12}. Similar to when using reproducing kernels, linearization is achieved by nonlinearly embedding or representing the input data in the state space of a reservoir system, which is the name given in this field to a randomly generated state-space system. It has been shown in the literature that, at least in the deterministic context, this technique has close ties with classical embedding strategies like Taken's Theorem~\cite{takensembedding} and generalized synchronizations~\cite{pecora:synch, ott2002chaos, boccaletti:reports:2002, eroglu2017synchronisation} (see~\cite{hart:ESNs, allen:tikhonov, RC18, RC21} for recent developments in that direction).

An explicit link between kernels and RC has been established in~\cite{RC16},  where an RKHS has been associated {with} any state-space system that satisfies the so-called echo state property~\cite{jaeger2001, Jaeger04} using the state functional~\cite{Boyd1985, RC6, RC7} as a feature map. We call the resulting kernel map the {\bfi reservoir kernel}. This idea will be presented in detail later on in Section~\ref{Setup and preliminary discussion} and lies also behind the developments  in~\cite{hermans2012recurrent, Tino2019}. There are two main difficulties linked to the use of reservoir kernels. First, computing them efficiently requires explicit analytic expressions for the state functional that are, in general, not available. Second, even when using {\it universal reservoir families} with linear readouts of the type studied in~\cite{RC6, RC7, RC8}, the {\it corresponding reservoir kernels are not necessarily universal} in the sense of~\cite{steinwart2001influence, micchelli2006universal}. More explicitly, universal reservoir families can be used to approximate arbitrarily well functionals in various categories. This means that given a functional, we can find an element in the universal reservoir family that approximates it arbitrarily well. However, in general, universal reservoir families do not (at least in a manageable way) produce a single kernel map whose kernel sections (in the terminology introduced in~\cite{micchelli2006universal}) are universal approximants. {\it This paper introduces a construction that simultaneously solves these two problems.}

\subsection{The Results}

After briefly stating our goals, we make the results of the paper more explicit. Our main objective is the construction of a universal reservoir kernel that can approximate any fading memory causal and time-invariant reservoir functional (as defined in~\cite{Boyd1985}, see also~\cite{manjunath:prsl, RC30, RC31, RC32} for recent work and characterizations). We shall call this the {\bfi fading memory property (FMP)} category.
Most theoretical developments in this direction are based on the possibility to represent any such filter, under certain restrictions on regularity and the inputs that we describe later on, as a {\bfi  Volterra series} expansion~\cite{wiener:book, sandberg1983series, Boyd1985, sandberg:time-delay}. The estimation of the coefficients of such a representation is an important question in signal processing and in engineering in general. Various recursive~\cite{poggio1975optimal}, kernel-based~\cite{dodd2002new, Franz2006}, or neural~\cite{wray1994calculation, de2019multi} methods have been proposed to tackle it. The strategies based on the use of kernels hinge on the construction of an RKHS using feature maps that contain all the monomials up to a certain order that appear in the Volterra series expansions, as well as in using the resulting kernel maps to solve the corresponding polynomial regressions. Implementing this specific approach in the FMP setup runs into two main challenges. The first one, already present when using kernel methods for arbitrary polynomial regressions, is to determine the highest polynomial order that is needed in a particular estimation task. The second one has to do with the FMP filter setup with semi-infinite inputs, in which one would have to determine how far into the past the inputs should be taken into account for a given Volterra series expansion. The approach that we propose tackles these two questions by proceeding in two steps that are contained in Section~\ref{The reservoir kernel of a Volterra series}:
\begin{description}
\item [(i)] We realize the Volterra series expansion of any FMP filter as a {\it linear functional of a {\bfi strongly universal} infinite-dimensional state-space system}. The term strongly universal means that the state-space system employed in this realization does not depend on the FMP filter that is being realized, and that only the linear functional that is used as readout needs to be adapted when different filters are considered. This situation is very close in philosophy to the approach followed in reservoir computing, and that is why we will refer to the infinite-dimensional system in this realization as the {\bfi  Volterra reservoir}.
\item [(ii)] Using the echo state property of the Volterra reservoir, we introduce the associated reservoir kernel map as in \eqref{definition kernel map} and we show that it can be readily computed using explicit recursions. This feature makes it possible in practice to implement the Volterra reservoir using the implicit reduction and kernelization procedures introduced in~\cite{RC16} (see Section~\ref{Setup and preliminary discussion} for details). We shall refer to the resulting kernel as the {\bfi  Volterra reservoir kernel}. The Volterra reservoir kernel  {\bfi is universal} in the sense of~\cite{steinwart2001influence, micchelli2006universal}, that is, its space of kernel sections coincides with the space of fading memory filters.
\end{description}
In Section~\ref{Numerical illustration}, we empirically assess the good performance and scalability of the Volterra reservoir kernel in a learning task associated with the modeling of the multidimensional covolatility of financial assets and compare it to other standard families of static and sequential kernels. The code and the data necessary to reproduce these numerical results are publicly available. We emphasize that the scope of the application of our Volterra kernel is very broad. It can be used as a general-purpose kernel for applied dynamic learning problems. These include the learning of filters (sequence-to-sequence maps) defined on sequence spaces, forecasting of time series, or the path-continuation of dynamical systems. The Volterra reservoir kernel has subsequently been used in~\cite{RC29} to construct an infinite-dimensional generalization of a polynomial approach to dynamic regressions called next-generation reservoir computing (NG-RC)~\cite{gauthier2021next}.

A first version of the approach in this paper has been presented in~\cite{RC13}, where a similar strategy was pursued for truncated Volterra series followed by random projections. Even though this strategy leads to important conclusions, such as the explanation of the expressive power of the randomly generated state-affine systems introduced in~\cite{RC6}, the truncations made impossible the recursive characterization of the reservoir kernel. 

The continuous-time counterpart of the results in this paper is much related to the {\bfi signature process}~\cite{chen1957integration, chen1958integration, Lyons:1998, hambly2010uniqueness, RC19} in rough path theory. This process allows for the representation of the solutions of affinely controlled differential equations as a linear functional of it, a feature shared, loosely speaking, by the solutions of the Volterra reservoir in the discrete-time setup. A kernel representation can also be constructed for the signature process~\cite{salvi2021signature}, which is the analog of the Volterra reservoir kernel and that, in this continuous-time setting, is characterized as the solution of a Goursat partial differential equation.

\subsection{Notation, setup, and preliminary discussion}
\label{Setup and preliminary discussion}

\medskip   \noindent {\bf Recurrent neural networks and reservoir computing.} We briefly introduce a few definitions that make explicit the setup in which we shall be working. The objects of interest in this paper are input/output systems determined by state-space systems. The symbols ${\cal Z}  $  and ${\cal Y} $ will denote the {\bfi input} and the {\bfi output spaces}, respectively, and ${\cal X} $ will be the {\bfi state space} of the system that will create the link between them. In Section~\ref{The reservoir kernel of a Volterra series} we will choose ${\cal Z}  = \mathbb{R} ^d$ and ${\cal Y}= \mathbb{R} ^m$, but the results stated in this section hold more generally. These three spaces are typically subsets of a Euclidean space or, more generally, finite or infinite-dimensional manifolds; for the time being, we shall assume no particular structure on them. A {\bfi  discrete-time state-space system}  is determined by the following two equations that put in relation sequences ${\bf z} \in {\cal Z}^{\mathbb{Z}}, {\bf y} \in {\cal Y}^{\mathbb{Z}}, {\bf x} \in {\cal X}^{\mathbb{Z}} $ in the three spaces that we just introduced:
\begin{empheq}[left={\empheqlbrace}]{align}
\mathbf{x} _t &=F(\mathbf{x} _{t-1}, {\bf z}_t),\label{rc state eq}\\
{\bf  y} _t &= h( \mathbf{x} _t) , \label{rc readout eq}
\end{empheq}
for any $t \in \Bbb Z $. The map $F:  {\cal X}  \times {\cal Z} \rightarrow {\cal X}$ is called the {\bfi  state map} and  $h:{\cal X} \rightarrow {\cal Y}$ the {\bfi  readout } or {\bfi  observation } map. We shall sometimes refer to such a system by using the triple $({\cal X}, F, h)$. The term {\bfi recurrent neural network (RNN)}  is sometimes used in the literature to refer to state-space systems where the state map $F$ in \eqref{rc state eq} is neural network-like, that is, it is a concatenation of compositions of a nonlinear activation function with an affine function of the states and the input. A particular case of RNNs are {\bfi echo state networks} introduced in~\cite{Matthews:thesis, Jaeger04} where one neural layer of this type is used (with random connectivity between neurons in~\cite{Jaeger04}). Cases in which the state map is randomly generated, at least in part, and the readout is functionally simple are collectively known as {\bfi reservoir computing (RC)} systems~\cite{jaeger2001, maass1, Jaeger:2002, Jaeger04, maass2} and have received much attention due to the ease in training that this approach entails and the universality approximation properties that it exhibits~\cite{RC6, RC7, RC8, RC10, RC12,RC24}.

We focus on state-space systems of type \eqref{rc state eq}-\eqref{rc readout eq} that determine an {\bfi input/output} system. This happens in the presence of the so-called {\bfi echo state property (ESP)}, that is, when for any  ${\bf z} \in {\cal Z}^{\mathbb{Z}}$ there exists a unique $\mathbf{y} \in {\cal Y}^{\mathbb{Z}}$ such that \eqref{rc state eq}-\eqref{rc readout eq} hold. In that case, the {\bfi  state-space filter}  $U ^F_h: {\cal Z}^{\mathbb{Z}}\rightarrow  {\cal Y}^{\mathbb{Z}} $ associated to the state-space system $({\cal X}, F, h)$  is well-defined by: 
\begin{equation*}
U  ^F _h({\bf z}):= \mathbf{y},
\end{equation*}
where ${\bf z} \in {\cal Z}^{\mathbb{Z}} $  and $\mathbf{y} \in {\cal Y}^{\mathbb{Z}} $ are linked by \eqref{rc state eq}-\eqref{rc readout eq} via the ESP. If the ESP holds at the level of the state equation \eqref{rc state eq}, we can define a {\bfi  state} or {\bfi  reservoir filter}  $U ^F: {\cal Z}^{\mathbb{Z}}\rightarrow  {\cal X}^{\mathbb{Z}} $ and, in that case, we have that
\begin{equation*}
U ^F_h:= h \circ U ^F. 
\end{equation*}
It is easy to show that state-space filters are automatically causal and time-invariant (see~\cite[Proposition 2.1]{RC7}) and hence it suffices to work with their restriction $U ^F_h: {\cal Z}^{\mathbb{Z}_-}\rightarrow  {\cal Y}^{\mathbb{Z}_-} $ to semi-infinite inputs and outputs, where we use the symbol $\mathbb{Z}_{-}  $ to denote the negative integers including zero. Moreover, $U ^F_h $  determines a state-space (also called reservoir) {\bfi   functional} $H ^F_h: {\cal Z}^{\mathbb{Z}_-}\rightarrow  {\cal Y}$ as $H ^F_h({\bf z}):=U ^F_h({\bf z})_0 $, for all ${\bf z} \in {\cal Z}^{\mathbb{Z}_{-}} $. The same applies to $U ^F $ and $H ^F $ when the ESP holds at the level of the state equation. If ${\cal Z} $ is a compact metric space and ${\cal Z}^{\mathbb{Z}_-}$ is endowed with the product topology, we say that $H ^F_h$ or $H ^F$ have the {\bfi fading memory property (FMP)} when they are continuous. This definition can be extended to unbounded input spaces using weighted norms (see~\cite{RC9} for a comprehensive treatment of this case as well as~\cite{manjunath:prsl, RC30, RC31, RC32} for a more general characterization of the FMP).
\medskip   

\noindent {\bf The kernel associated with a state-space system.} In the following paragraphs, we briefly recall the construction introduced in~\cite{RC16}  in which given a state-space system $F:  {\cal X}  \times {\cal Z} \longrightarrow {\cal X}$ that satisfies the echo state property, a kernel map on the input sequences space was defined using the state functional $H ^F: {\cal Z}^{\Bbb Z_-}\longrightarrow {\cal X}  $ as a feature map. 

The importance of this construction is in the fact that it allows us, using the classical Representer Theorem~\cite[page 117]{Mohri:learning:2012}, to reduce the search for a linear readout which is optimal with respect to the regularized empirical risk minimization associated to any loss to the search for a readout defined on the smaller space ${\cal X} _R$ 
given by the closure of the linear span (in the Hilbert space ${\cal X}  $) of the set of reachable states, that is,
\begin{equation}
\label{linear span def}
{\cal X} _R:= \overline{{\rm span} \left\{{\cal X} _R\right\}}= \overline{{\rm span}\left\{H ^F({\bf z})\mid {\bf z}\in  {\cal Z}^{\Bbb Z _-}\right\}}. 
\end{equation}
This approach allows one to carry out dimension reduction without actually having to compute the set of reachable states. This procedure was called in~\cite{RC16}  {\bfi implicit reduction}.

Let $F:  {\cal X}  \times {\cal Z} \longrightarrow {\cal X}$ be a state map such that the pair $\left({\cal X}, \langle \cdot , \cdot \rangle_{{\cal X}}\right) $ is a Hilbert space and $F$ has the echo state property {as well as the fading memory property, that is, the corresponding state functional $H ^F: {\cal Z}^{\Bbb Z_-}\longrightarrow {\cal X}  $ is continuous}. Define the {\bfi reservoir  kernel map} 
\begin{equation}
\label{definition kernel map}
\begin{array}{cccc}
K:&{\cal Z}^{\Bbb Z_-} \times {\cal Z}^{\Bbb Z_-} &\longrightarrow & \mathbb{R}\\
	&({\bf z}, {\bf z}')&\longmapsto & \langle H ^F({\bf z}) , H ^F({\bf z}') \rangle_{{\cal X}}.
\end{array}
\end{equation}
$K$ is hence a kernel map that has been constructed using the reservoir functional $H ^F: {\cal Z}^{\Bbb Z_-} \longrightarrow {\cal X} $  as a feature map.
The reservoir kernel $K$ is obviously symmetric and positive semidefinite in the sense that for any $a _i \in \mathbb{R} $, ${\bf z} _i \in {\cal Z}^{\Bbb Z_-} $, $i \in \left\{1, \ldots, n\right\} $, we have that $\sum_{i,j=1}^n a _i a _jK({\bf z} _i, {\bf z} _j)\geq 0 $. {Moreover, as the functional $H ^F $ is continuous by the FMP then so is the kernel $K$.} Let $\left(\mathbb{H}, \langle \cdot , \cdot \rangle_{\mathbb{H}}\right)$ be the corresponding RKHS given by
\begin{equation}
\label{definition H for F}
\mathbb{H}:= \overline{{\rm span}\left\{K _{\bf z}:=K({\bf z}, \cdot ):{\cal Z}^{\Bbb Z_-} \longrightarrow \mathbb{R}\mid {\bf z} \in {\cal Z}^{\Bbb Z_-} \right\}}
\end{equation}
made out of finite linear combinations of elements of the type $K _{\bf z} (\cdot )=K({\bf z}, \cdot )= \langle H ^F({\bf z}) , H ^F(\cdot ) \rangle_{{\cal X}}$, ${\bf z} \in {\cal Z}^{\Bbb Z_-} $, together with all the limits of Cauchy sequences with respect to the metric induced by the inner product obtained as the linear extension of 
$
\langle K _{\bf z}, K _{{\bf z}'}\rangle_{\mathbb{H}}=K({\bf z}, {\bf z}')=\langle H ^F({\bf z}) , H ^F({\bf z}') \rangle_{{\cal X}}$, ${\bf z}, {\bf z}' \in {\cal Z}^{{\Bbb Z}_-}$. 
Notice now that since ${\cal X} _R $ is a Hilbert subspace of ${\cal X} $ then both $H ^F: {\cal Z}^{\Bbb Z_-}\longrightarrow {\cal X}$ and its analog (denoted with the same symbol) $H ^F: {\cal Z}^{\Bbb Z_-}\longrightarrow {\cal X} _R$ obtained by restricting the codomain are, by construction, feature maps of the kernel $K$. Using this fact, it can be shown (cf.~\cite[Theorem 4.21]{Steinwart2008}) that 
\begin{equation}
\label{first identity H}
\mathbb{H}= \left\{\langle {\bf W}, H ^F(\cdot )\rangle_{{\cal X}}\mid {\bf W} \in {\cal X} _R\right\}.
\end{equation}

The use of the kernel \eqref{definition kernel map} and the corresponding RKHS \eqref{definition H for F}-\eqref{first identity H} can prove very advantageous when using systems of the form \eqref{rc state eq}-\eqref{rc readout eq} in practice and where the readout is linear (that is, in \eqref{rc readout eq} one has $h( \mathbf{x})= \left\langle {\bf W},  \mathbf{x}\right\rangle_{{\cal X}}$, for some ${\bf W} \in {\cal X} $). A common estimation problem in that case is finding the readout vector ${\bf W} \in {\cal X} $ that minimizes the empirical risk associated with a prescribed loss function with respect to a finite sample of input/output observations. How to approach that problem in our dynamic setup using the kernel $K:{\cal Z}^{\Bbb Z_-} \times {\cal Z}^{\Bbb Z_-} \longrightarrow  \mathbb{R}$ is explained in Appendix~\ref{Kernelization Appendix}. The main difficulty in all the kernel-based methodologies in our case is that the kernel needs to be evaluated on the available data to construct the so-called Gram matrices (see Appendix~\ref{Kernelization Appendix}). This may be problematic for nonlinear state-space systems since, in general, an explicit expression of the functional $H ^F $ is not available. In the next section, this issue is addressed for a particular but important case, the so-called Volterra kernel, via a method that we call {\bfi sequentialization} and that allows the computation of the kernel recurrently.

\section{The reservoir kernel of a Volterra series}
\label{The reservoir kernel of a Volterra series}

All along this section, we consider as input  and output spaces ${\cal Z}  $ and ${\cal Y}  $ two finite-dimensional Euclidean spaces of dimension $d$ and $m$, respectively, that is, ${\cal Z}  = \mathbb{R} ^d$ and ${\cal Y}= \mathbb{R} ^m $. We denote by $\langle \cdot , \cdot \rangle $ and $\|{\cdot}\|$   the Euclidean inner product and norm, {as well as certain inner products and norms induced by it}. Additionally, for  any ${\bf z} \in (\mathbb{R}^d)^{\mathbb{Z}_{-}}$ we define the $p$-norms as $\left\|{\bf z}\right\|_ p := \left( \sum_{t\in \mathbb{Z}_-} \|{\bf z}_t \|^p \right)^{1/p}  $,  for $1 \le p <\infty $, and  $\left\|{\bf z}\right\|_ \infty:=\sup_{t \in \mathbb{Z}_{-}}  \left\{\left\|{\bf z} _t\right\|\right\}$ for $p=\infty$. Given $M>0 $, we denote by $K _M:= \{{\bf z} \in (\mathbb{R}^d)^{\mathbb{Z}_{-}}\mid \left\|{\bf z} _t\right\|\leq M \enspace {\rm for} \enspace {\rm all} \enspace t\in \Bbb Z _-\}$. It is easy to see that $K_M = \overline{B _M}
\subset \ell_{-}^{\infty}(\mathbb{R}^d)$, with 
$
B _M:=B_{\left\|\cdot \right\|_\infty}({\bf 0}, M)$  and $\ell_{-}^{\infty}(\mathbb{R}^d):= \{{\bf z} \in (\mathbb{R}^d)^{\mathbb{Z}_{-}}\mid \left\|{\bf z}\right\|_ \infty< \infty \}$. {{We define $\widetilde{B}_M:=B_M\cap \ell_{-}^{1}(\mathbb{R}^d)$ with $\ell_{-}^{1}(\mathbb{R}^d):= \{{\bf z} \in (\mathbb{R}^d)^{\mathbb{Z}_{-}}\mid \left\|{\bf z}\right\|_ 1< \infty \}$.} Additionally, we will write $L(V,W)$ to refer to the space of linear maps between the real vector spaces $V$ and $W$}.  

\subsection{The Volterra reservoir}
\medskip   \noindent {\bf Volterra series for FMP filters.}
We first recall that in a setup in which the inputs and outputs of a causal and time-invariant filter $U:K _M \subset \ell_{-}^{\infty}(\mathbb{R}^d) \rightarrow K _L \subset \ell_{-}^{\infty}(\mathbb{R}^m) $, $M, L>0 $,  are uniformly bounded, the fading memory property is just a continuity requirement on the map $U$ when $K _M$  and $K _L $ are endowed with the product topology~\cite[Proposition 2.11]{RC7}. The following theorem is a straightforward multivariate generalization of Theorem 29 in~\cite{RC9}. 
\begin{theorem}[Volterra representation of FMP filters]
\label{volterra representation}
Let $M, L>0 $ and let $U:K _M \subset \ell_{-}^{\infty}(\mathbb{R}^d) \rightarrow K _L \subset \ell_{-}^{\infty}(\mathbb{R}^m) $ be a causal and time-invariant fading memory filter whose restriction $U|_{B _M}$ is analytic as a map between open sets in the Banach spaces $\ell_{-}^{\infty}(\mathbb{R}^d) $ and  $\ell_{-}^{\infty}(\mathbb{R}^m)$ and satisfies $U({\bf 0})= {\bf 0} $. Then there exists a Volterra series representation of $U$ given {for any ${\bf z}\in\widetilde{B}_M$} by
\begin{equation}
\label{volterra representation eq}
\footnotesize
U ({\bf z})_t=\sum_{j=1}^{\infty}\sum_{m _1=- \infty}^0 \cdots\sum_{m _j=- \infty}^0 g _j(m _1, \ldots, m _j)({\bf z}_{ m _1+ t}\otimes \cdots \otimes{\bf z}_{ m _j+ t})
\end{equation}
with $ t \in \mathbb{Z}_{-}$ and where the map $g _j: (\Bbb Z_-) ^j \rightarrow L({\Bbb R}^d \otimes \cdots \otimes{\Bbb R}^d, {\Bbb R}^m) $ is given by
\begin{equation*}
\label{definition g maps}
g _j(m _1, \ldots, m _j)({\rm \mathbf{e}}_{i _1}\otimes \cdots \otimes{\rm \mathbf{e}}_{i _j})= \frac{1}{j!}D ^jH _U({\bf 0})({\rm \mathbf{e}}^{i _1}_{m _1}, \ldots ,{\rm \mathbf{e}}^{i _j}_{m _j}),
\end{equation*}
where, {for any ${\bf{z}}_0 $ in some open subset of $\ell_{-}^{\infty}(\mathbb{R}^d) $, $D ^jH _U({\bf{z}}_0)$ with $j\ge 1$ denotes the $j$-order Fr\'echet differential at ${\bf{z}}_0 $ of the functional $H _U $ associated to the filter $U$},  $\left\{\mathbf{e}_1, \ldots, \mathbf{e}_d\right\}$ is the canonical basis of $\mathbb{R}^d  $ and the sequences ${\rm \mathbf{e}}^{i _l}_{m _k}\in \ell_{-}^{\infty}(\mathbb{R}^d)$ are defined by:
\begin{equation*}
\label{canonical vectors sequences}
\left({\rm \mathbf{e}}^{i _l}_{m _k}\right)_t:=
\left\{
\begin{array}{cr}
\mathbf{e}_{i _l}\in {\Bbb R}^d, &{\rm \it if}\ t=m _k,\\
{\bf 0},& {\rm \it otherwise.}
\end{array}
\right.
\end{equation*}
\end{theorem}

\medskip   \noindent {\bf State-space representation of Volterra series.} We now show that there exists a state-space system with states in an infinite-dimensional tensor space such that any Volterra filter as in \eqref{volterra representation eq} can be realized as a linear readout of that system. To introduce the state space, we recall some tensor constructions. First, for any $l,d \in \mathbb{N} $, we denote by $T ^l( \mathbb{R} ^d)$ the space of tensors of order $l$ on $\mathbb{R} ^d $, that is,
\begin{equation*}
T ^l( \mathbb{R} ^d):= \bigg\{\sum_{i _1, \ldots, i _l=1}^d a_{i _1, \ldots, i _l}\mathbf{e}_{i _1}\otimes \cdots \otimes\mathbf{e}_{i _l}\mid a_{i _1, \ldots, i _l} \in \mathbb{R}\bigg\},
\end{equation*}
where $\left\{\mathbf{e}_{i }\right\}_{i \in \left\{1, \ldots, d\right\}} $ is the canonical basis in ${\Bbb R}^d $. By convention, one defines $T ^0( \mathbb{R} ^d)= \mathbb{R} $. The {\bfi  tensor algebra} $T ( \mathbb{R} ^d) $ of ${\Bbb R}^d  $ and the space of {\bfi  tensor series} $T(( \mathbb{R} ^d))$ are defined as the external direct sum and external direct product of the tensor spaces of all orders, respectively, that is,
\begin{equation*}
\begin{aligned}
T ( \mathbb{R} ^d) &=\bigoplus _{l=0} ^{\infty}T ^l( \mathbb{R} ^d),
\\ T (( \mathbb{R} ^d)) & =\prod_{l=0} ^{\infty}T ^l( \mathbb{R} ^d):= 
\bigg\{\sum_{l=0}^{\infty} \mathbf{v} _l \mid \mathbf{v} _l \in T ^l( \mathbb{R} ^d)\bigg\}\\
=\bigg\{\sum_{l=0}^{\infty} &\sum_{i _1, \ldots, i _l=1}^d a_{i _1, \ldots, i _l}\mathbf{e}_{i _1}\otimes \cdots \otimes\mathbf{e}_{i _l}\mid a_{i _1, \ldots, i _l} \in \mathbb{R}\bigg\},
\end{aligned}
\end{equation*}
where we emphasize that the sum going to infinity is external, while the sum going to $d $  is internal in $T ^l( \mathbb{R} ^d) $. 
The tensor algebra $T( \mathbb{R} ^d)$ consists of those elements of $T (( \mathbb{R} ^d))$ with only finitely many non-zero coefficients. Both $T^l(\mathbb{R}^d)$ and $T(\mathbb{R}^d)$ naturally inherit an inner product structure from the Euclidean inner product. { For any $\mathbf{v} _l, \mathbf{w} _l \in T ^l( \mathbb{R} ^d)$, $l \in \mathbb{N} $ the inner product is explicitly given by}
\begin{equation}
\label{dif inner prod prod}
\left\langle\sum_{l=0}^{\infty} \mathbf{v} _l , \sum_{l=0}^{\infty} \mathbf{w} _l  \right\rangle=\sum_{l=0}^{\infty}\left\langle \mathbf{v} _l ,  \mathbf{w} _l  \right\rangle.
\end{equation}
Using the expression of the elements in the tensor algebra in terms of the canonical basis in ${\Bbb R}^d $, this amounts to: 
\tiny
\begin{multline*}
\!\!\!\!\!\!\!\!\left\langle \sum_{l=0}^{\infty}\sum_{i _1, \ldots, i _l=1}^d a_{i _1, \ldots, i _l}\mathbf{e}_{i _1}\otimes \cdots \otimes\mathbf{e}_{i _l}, \sum_{l=0}^{\infty}\sum_{i _1, \ldots, i _l=1}^d b_{i _1, \ldots, i _l}\mathbf{e}_{i _1}\otimes \cdots \otimes\mathbf{e}_{i _l}   \right\rangle\\=\sum_{l=0}^{\infty}\sum_{i _1, \ldots, i _l=1}^d a_{i _1, \ldots, i _l}b_{i _1, \ldots, i _l}.
\end{multline*}
\normalsize
Denote by $\| \cdot \|$ the norm induced on $T(\mathbb{R}^d)$ by this inner product and let $\overline{T(\mathbb{R}^d)}$ be the completion  of  $T(\mathbb{R}^d)$ by this norm. Then  $\overline{T(\mathbb{R}^d)}$ is a Hilbert space consisting precisely of the series in $T((\mathbb{R}^d))$ with finite norm.

The procedure that we just used to construct the completed tensor algebra $\overline{T(\mathbb{R}^d)}$ out of the vector space ${\Bbb R}^d $ can be repeated to obtain the completed tensor algebra  $\overline{T(\overline{T( \mathbb{R} ^d)} )}$ of $\overline{T( \mathbb{R} ^d)}$. We will refer to  $\overline{T(\overline{T( \mathbb{R} ^d)} )}$ as the {\bfi  double tensor algebra} of  ${\Bbb R}^d  $ and we will denote it by $T_2( \mathbb{R} ^d)= \overline{T(\overline{T( \mathbb{R} ^d)} )}$. We emphasize that since $\overline{T( \mathbb{R} ^d)}  $ is an inner product space, then using again \eqref{dif inner prod prod} so is $T_2( \mathbb{R} ^d) $.

Before we introduce the Volterra reservoir, we state a last definition. Given $\tau \in \mathbb{R}  $ and $ {\bf z} \in {\Bbb R}^d  $, we define its $\tau$-{\bfi tensorization} $\widetilde{{\bf z}} \in T((\mathbb{R}^d)) $  as
$
{\displaystyle \widetilde{{\bf z}} =1+\sum _{i=1} ^{\infty} \tau^i \underbrace{{\bf z}\otimes \cdots \otimes {\bf z}}_\text{$i$-times}}
$. Using the norm in $T((\mathbb{R}^d))$ associated to the inner product introduced in \eqref{dif inner prod prod}, it is easy to see that
\begin{equation}
\label{norm of tilde}
{\displaystyle \left\|\widetilde{{\bf z}}\right\| ^2= 1/ \left(1- \tau ^2 \left\|{\bf z}\right\|^2\right)},\quad \mbox{ whenever $\tau^2  \left\|{\bf z}\right\|^2<1 $}
\end{equation}
and so $\widetilde{{\bf z}} \in \overline{T( \mathbb{R} ^d)}$ in that case.

\begin{theorem}[The Volterra reservoir]
\label{The Volterra reservoir}
Let $M>0 $ and let $\tau>0 $ be such that $\tau^2  M^2<1 $. Let $0< \lambda< \sqrt{1- \tau ^2 M ^2} $. Consider now the state system with uniformly bounded inputs in $K _M \subset \ell_{-}^{\infty}(\mathbb{R}^d) $ and states in the double tensor algebra $T_2( \mathbb{R} ^d)$ given by the recursion
\begin{equation}
\label{Volterra reservoir eq}
\mathbf{x} _t= \lambda \mathbf{x} _{t-1} \otimes \widetilde{{\bf z}} _t+ 1, \quad\mbox{$t \in \mathbb{Z}_{-} $}
\end{equation}
where $\widetilde{{\bf z}} _t \in \overline{T( \mathbb{R} ^d)} $ is the $\tau $-tensorization of ${\bf z} _t  $. Then,
\begin{description}
\item [(i)] This system has the ESP and hence defines a unique causal, time-invariant, and FMP filter $U ^{{\rm Volt}} _\lambda:K _M \subset \ell_{-}^{\infty}(\mathbb{R}^d) \rightarrow \{{\bf x} \in (T_2( \mathbb{R} ^d))^{\mathbb{Z}_{-}}\mid \left\|{\bf x} _t\right\|\leq L \enspace {\rm for} \enspace {\rm all} \enspace t\in \Bbb Z _-\}$ given by {
\begin{equation}
\label{solution uvolt}
U ^{{\rm Volt}}  _\lambda({\bf z})_t=1+\sum_{j=0}^{\infty} \lambda^{j+1} \bigotimes_{k=0}^{j} \widetilde{{\bf z}} _{t-k},\enspace\mbox{$t \in \mathbb{Z}_{-} $}
\end{equation}}
with $L=1/ \left(1- \lambda/\sqrt{1- \tau ^2M ^2}\right)  $.
\item [(ii)] Let $U:K _M \subset \ell_{-}^{\infty}(\mathbb{R}^d) \rightarrow K _L \subset \ell_{-}^{\infty}(\mathbb{R}^m) $ be a causal, time-invariant, FMP filter whose restriction $U|_{{B} _M}$ is analytic as a map between open sets in the Banach spaces $\ell_{-}^{\infty}(\mathbb{R}^d) $ and  $\ell_{-}^{\infty}(\mathbb{R}^m)$ and satisfies $U({\bf 0})= {\bf 0} $. Then, there exists a unique linear map ${\bf W} \in L(T _2({\Bbb R}^d), {\Bbb R}^m)$, such that
{
\begin{equation}
\label{linear funct of volterra}
U ( {\bf z})_t={\bf W}U ^{{\rm Volt}}  _\lambda({\bf z})_t, \enspace  \forall {\bf z}\in\widetilde{B}_M, \enspace t \in \mathbb{Z}_-.
\end{equation}}
We shall refer  to the map $F^{{\rm Volt}}_\lambda:T_2( \mathbb{R} ^d) \times \{{\bf z} \in  \mathbb{R} ^d \mid \left\|{\bf z}\right\|\leq M \} \rightarrow T_2( \mathbb{R} ^d) $ given by $F^{{\rm Volt}}_\lambda (\mathbf{x} ,  {\bf z} ):=  \lambda \mathbf{x}  \otimes \widetilde{{\bf z}} + 1 $ as the {\bfi  Volterra reservoir map}, to the filter $U ^{{\rm Volt}}  _\lambda $ in \eqref{solution uvolt} as the {\bfi  Volterra filter}, and to the equality \eqref{linear funct of volterra} as the {\bfi  Volterra filter representation} of the FMP filter $U$. The corresponding functional $H ^{{\rm Volt}}  _\lambda ({\bf z}):=U ^{{\rm Volt}}  _\lambda({\bf z}) _0$ will be called the {\bfi  Volterra functional}.
\end{description}
\end{theorem}
\noindent {The proof of Theorem~\ref{The Volterra reservoir} is provided in Appendix~\ref{Appendix 1}. }

\begin{remark}
\label{strong universality volterra}
\normalfont
The Volterra filter representation \eqref{linear funct of volterra} proves the {\bfi strong universality} property of the Volterra reservoir that we announced in the introduction. This means that in \eqref{linear funct of volterra}, the term $U ^{{\rm Volt}}  _\lambda({\bf z}) $ depends only on the input ${\bf z}  $ and not on the FMP filter $U$ that is being represented. To adjust the representation \eqref{linear funct of volterra} for another filter $U'$, it suffices to replace the linear readout ${\bf W}$ by a new readout ${\bf W}'$.
\end{remark}

\subsection{The Volterra reservoir kernel}

Theorem~\ref{The Volterra reservoir} shows two important features of the Volterra reservoir. On the one hand,  the equality \eqref{linear funct of volterra} proves that any FMP filter can be learnt by estimating a linear readout of the Volterra filter, even though the state space is, in this case, infinite-dimensional. Nevertheless, as we show in the following result, the associated reservoir kernel map admits an explicit expression that can be sequentially computed and that allows kernelizing the estimation problem and reducing it to a regression problem in the span of the data (see Proposition~\ref{Implicit reduction and kernelization} in Appendix~\ref{Kernelization Appendix}).

Before we state the proposition, we introduce two definitions. Let $\tau \in \mathbb{Z}_{-} $ and define the $\tau${\bfi  -time delay operator} $T _{-\tau}: (\mathbb{R}^d)^{\Bbb Z _-} \rightarrow (\mathbb{R}^d)^{\Bbb Z _-} $  by  $T_{-\tau}( {\bf z}) _t:= {\bf z}_{t+ \tau}$,  for any ${\bf z} \in (\mathbb{R}^d)^{\Bbb Z _-}  $ and $t \in \mathbb{Z}_{-} $. Additionally, for any $t \in \Bbb Z_- $, we denote by $p _t:(\mathbb{R}^d)^{\Bbb Z _-} \rightarrow {\Bbb R}^d $ the $t$-{\bfi  projection} given by $p _t({\bf z})= {\bf z} _t $. Finally, recall that a filter $U: (\mathbb{R}^d)^{\Bbb Z _-}  \rightarrow (\mathbb{R}^m)^{\Bbb Z _-}  $ is time-invariant when $U \circ T _{- \tau}= T _{- \tau} \circ U $, for any $\tau \in \mathbb{Z}_{-} $.

\begin{proposition}[The Volterra reservoir kernel]
\label{The Volterra reservoir kernel proposition}
Let $M>0 $ and let $\tau>0 $ be such that $\tau^2  M^2<1 $. Let $0< \lambda< \sqrt{1- \tau ^2 M ^2} $. The reservoir kernel map $K^{{\rm Volt}}:K _M \times K _M \rightarrow \mathbb{R} $  of the Volterra reservoir in \eqref{Volterra reservoir eq} defined using the parameter $\lambda$, $\tau$-tensorization, and with uniformly bounded inputs in $K _M \subset \ell_{-}^{\infty}(\mathbb{R}^d) $ is well-defined and can be explicitly calculated for ${\bf z}, {\bf z}' \in K _M$ using the recursion
\begin{equation}
\label{recursion for kvolt}
K^{{\rm Volt}}({\bf z}, {\bf z}')=1+ \frac{\lambda ^2  K^{{\rm Volt}}(T _1({\bf z}), T _1( {\bf z}'))}{1- \tau ^2\langle p _0({\bf z}), p _0({\bf z}') \rangle},
\end{equation}
or the expression of its unique solution
\begin{equation}
\label{recursion for kvolt sum}
{ K^{{\rm Volt}}({\bf z}, {\bf z}')
=1+ \sum_{k=1} ^{\infty} \lambda^{2k} \prod_{j=0}^{k-1} \frac{1  }{1- \tau ^2\langle {\bf z}_{-j}, {\bf z}_{-j}' \rangle}.}
\end{equation}
\end{proposition}
\noindent {The proof of Proposition~\ref{The Volterra reservoir kernel proposition} is provided in Appendix~\ref{Appendix 2}. }

\medskip

\noindent {\bf The recursion for the computation of the Volterra Gram matrix.} Expressions \eqref{recursion for kvolt} and \eqref{recursion for kvolt sum} can be used to compute the Gram matrix \eqref{gram matrix definition} ${\mathcal K}^{{\rm Volt}} $ that one can associate to the Volterra reservoir kernel $K^{{\rm Volt}} $ when dealing with a finite-length input $\{{\bf Z}_{-i}\}_{i\in \left\{ 0, \dots, n-1\right\}}$ of size $n$. As we did in \eqref{finite sample}, for each time step $i\in \{ 0, \dots, n-1\}$ we define the truncated input samples as
$
{\bf Z}_{-i}^{-n+1} := (\ldots,{\bf 0},{\bf 0},{\bf Z}_{-n+1},\ldots,{\bf Z}_{-i-1},{\bf Z}_{-i})$, and we then define the Gram matrix ${\mathcal K}^{{\rm Volt}} \in \mathbb{M} _n $ of the sample  as
\begin{equation*}
{\mathcal K}^{{\rm Volt}} _{i,j}=K^{{\rm Volt}}\left({\bf Z}_{-(n-i)}^{-n+1}, {\bf Z}_{-(n-j)}^{-n+1}\right), \quad \mbox{$i,j \in \left\{1, \ldots, n\right\}$}.
\end{equation*}
These entries can be recursively computed using \eqref{recursion for kvolt} by setting ${\mathcal K}^{{\rm Volt}} _{0,0}={\mathcal K}^{{\rm Volt}} _{i,0}=1 / (1- \lambda ^2)$ for all $i\in \left\{1, \ldots, n\right\}$ and by computing
\begin{equation}
\label{recursion gramian}
{\mathcal K}^{{\rm Volt}} _{i,j}=1 + \lambda ^2 \frac{{\mathcal K}^{{\rm Volt}} _{i-1,j-1}}{1- \tau ^2 \left\langle {\bf Z}_{-(n-i)},  {\bf Z}_{-(n-j)}\right\rangle}
\end{equation}
for all $j \in  \left\{1, \ldots, i\right\}$.
This set of equations defines the lower triangular part of the Gram matrix, which can be completed by using its symmetry. We note that the solution obtained by using \eqref{recursion gramian} does depend on the initial conditions ${\mathcal K}^{{\rm Volt}} _{0,0}={\mathcal K}^{{\rm Volt}} _{i,0}=1/ (1- \lambda ^2) $, $i\in \left\{1, \ldots, n\right\}$. These choices are consistent with the completion using zeros that was used in \eqref{finite sample} to construct a semi-infinite vector out of the finite sample and yield the exact solution \eqref{recursion for kvolt sum} for this choice of completion.  Nevertheless, the contractive character of \eqref{recursion for kvolt} makes the resulting values asymptotically independent of the initial condition, that is, the entries ${\mathcal K}^{{\rm Volt}} _{i,j} $ of the Gram matrix as computed in \eqref{recursion gramian} depend increasingly less on the initial condition as $i$  and $j$ become higher. This asymptotic independence is more pronounced as the parameter $\lambda$ is chosen smaller.

\medskip   \noindent {\bf Forecasting using the Volterra kernel.}  In the presence of a sample of input/output observations $\{( {\bf Z}_{-i}, {Y}_{-i})\}_{i\in \left\{ 0, \dots, n-1\right\}}$, the Gram matrix computed in \eqref{recursion gramian} can be used to solve the optimization problem \eqref{optimization in terms of gramian}. Its solution yields a set  $\alpha _1, \ldots, \alpha_{n} \in \mathbb{R}$ which allows us to explicitly write the best estimate of the input/output map using empirical risk minimization, namely,
\begin{align*}
H^{F^{{\rm Volt}}_\lambda}_{\widehat{{\bf W}}} (\cdot )&=
\sum_{i=1}^{n} \alpha _iK^{{\rm Volt}}_{{\bf Z}_{-(n-i)}^{-n+1}}(\cdot )\\
&= \left\langle 
\sum_{i=1}^{n} \alpha _iH^{{\rm Volt}}_\lambda \left({\bf Z}_{-(n-i)}^{-n+1}\right), H^{{\rm Volt}}_\lambda(\cdot )
\right\rangle,
\end{align*}
where $\widehat{{\bf W}}=\sum_{i=1}^{n} \alpha _iH^{{\rm Volt}}_\lambda \left({\bf Z}_{-(n-i)}^{-n+1}\right) $.

If we now have available a new set of inputs $ \left\{ {\bf Z}_1, \ldots, {\bf Z} _h\right\} $, this estimated functional can be used to forecast the outputs   $ \left\{ \widehat{{Y}}_1, \ldots, \widehat{Y} _h\right\} $ of the data generating process by computing
\begin{equation}
\label{forecast with alphas}
\widehat{Y }_j= {H^{F^{{\rm Volt}}_\lambda}_{\widehat{{\bf W}}} }\left({\bf Z}_{j}^{-n+1}\right)= \sum_{i=1}^{n} \alpha _iK^{{\rm Volt}}\left({{\bf Z}_{-(n-i)}^{-n+1}}, {\bf Z}_{j}^{-n+1}\right)
\end{equation}
for all $j \in \left\{1, \ldots, h\right\}$, and where $${\bf Z}_{j}^{-n+1}=(\ldots,{\bf 0},{\bf 0},{\bf Z}_{-n+1},\ldots,{\bf Z}_{-1},{\bf Z}_{0}, {\bf Z}_{1}, \ldots, {\bf Z} _j).$$ 
The evaluations of the Volterra kernel in \eqref{forecast with alphas} can be written down in terms of an extension of the Gram matrix ${\mathcal K}^{{\rm Volt}} \in \mathbb{M} _n  $ that we computed in \eqref{recursion gramian} to a rectangular matrix ${\mathcal K}^{{\rm Volt}} \in \mathbb{M} _{n, n+h}  $ by using the recursions
\begin{equation*}
\label{recursion gramian extended}
{\mathcal K}^{{\rm Volt}} _{i,n+j}=1 + \lambda ^2 \frac{{\mathcal K}^{{\rm Volt}} _{i-1,n+j-1}}{1- \tau ^2 \left\langle {\bf Z}_{-(n-i)},  {\bf Z}_{j}\right\rangle}, \quad \mbox{for all $j \in  \left\{1, \ldots, h\right\}$},
\end{equation*}
that can be initialized by setting ${\mathcal K}^{{\rm Volt}} _{0,n}={\mathcal K}^{{\rm Volt}} _{0,n+1}= \cdots ={\mathcal K}^{{\rm Volt}} _{0,n+h}=1 / (1- \lambda ^2)$. The forecast \eqref{forecast with alphas} is then written in terms of these Gram matrix entries as
\begin{equation*}
\label{forecast with alphas with gram}
\widehat{Y }_j= {H^{F^{{\rm Volt}}_\lambda}_{\widehat{{\bf W}}} }\left({\bf Z}_{j}^{-n+1}\right)= \sum_{i=1}^{n} \alpha _i{\mathcal K}^{{\rm Volt}} _{i,n+j}
\end{equation*}
for all $j \in \left\{1, \ldots, h\right\}$.

\subsection{The Volterra kernel is universal}

We now explain how the strong universality of the Volterra reservoir that we commented on in Remark~\ref{strong universality volterra} is related to the universality of its kernel $K^{{\rm Volt}} $ in the classical sense introduced in~\cite{steinwart2001influence, micchelli2006universal} that we recall in Appendix~\ref{Kernel universality}. In that remark and in the theorem that precedes it, we showed that analytic filters with absolutely summable inputs can be {\it uniquely represented} using Volterra filters. The universality of the Volterra reservoir kernel that we establish in the next theorem yields an approximation rather than a representation result for a much larger class of filters and inputs. More specifically, this universality result means that any fading memory filter with uniformly bounded inputs can be {\it uniformly approximated by elements in the RKHS} generated by the Volterra kernel.

\begin{theorem}[Universality of the Volterra reservoir kernel]
	\label{thm:universality}
Let $K^{{\rm Volt}}:K _M \times K _M \longrightarrow \mathbb{R} $ be the reservoir kernel map introduced in Proposition~\ref{The Volterra reservoir kernel proposition} and let $K ^{{\rm Volt}} (K _M)$ be the associated space of kernel sections. Then
\begin{equation}
\label{kvolt is universal}
K ^{{\rm Volt}} (K _M)=C ^0(K _M)
\end{equation}
that is, the {\bfi  Volterra reservoir kernel is universal}.
\end{theorem}

{The proof of Theorem~\ref{thm:universality} is provided in Appendix~\ref{Appendix 3} below.}

\section{Numerical illustration}
\label{Numerical illustration}

{We now illustrate the empirical performance of the Volterra reservoir kernel in the learning of the Input/Output (IO) system given by the so-called BEKK model~\cite{engle:bekk}. BEKK is a parametric discrete-time time series model that is used in the financial econometrics literature and by practitioners for the modeling and forecasting of the conditional covariances of the returns of stocks traded in financial markets. We consider $d$ assets and the  BEKK($1$,$0$,$1$) model~\cite{engle:bekk} for their log-returns $\mathbf{r}_t \in \mathbb{R}^d$, $t\in \mathbb{Z}$, and associated conditional covariances $H_t = \text{Cov}(\mathbf{r}_t, \mathbf{r}_t|\mathbf{r}_{t-1},  \mathbf{r}_{t-2}, \ldots) \in \mathbb{S}_d$, $t\in \mathbb{Z}$:
\begin{align}
	\mathbf{r}_t &= H_t^{1/2} \mathbf{z}_t, \enspace \mathbf{z}_t \sim \text{IIDN}(\mathbf{0}_d, \mathbb{I}_d),\label{ret}\\
	H_t &= CC^\top + A \mathbf{r}_{t-1}\mathbf{r}_{t-1}^\top A^\top + B H_{t-1}B^\top,\label{H}
\end{align}
for which there exists a unique stationary and ergodic solution whenever $\rho(A \otimes A +B \otimes B)<1$~\cite{Boussama2011}.
For our empirical learning task, we consider the diagonal BEKK specification, where $C$ is an upper triangular matrix, $A$ and $B$ are diagonal matrices of dimension $d$, for which the above condition reduces to  $A_{ii}>0$, $|B_{ii}|<1$ for $i=1, \ldots, d$.  In our empirical learning task, we shall consider the IO system associated with this model in which the inputs are the Gaussian IID input innovations $\{\mathbf{z}_t\}_{t\in \mathbb{Z}}$ and the output observations are the conditional covariances $\{H_t\}_{t\in \mathbb{Z}}$, which are a highly nonlinear function of the input history. Since the covariance matrices are symmetric, we are interested only in learning the outputs $\{\mathbf{h}_t\}_{t\in \mathbb{Z}}$ with $\mathbf{h}_t = {\rm vech}(H_t) \in \mathbb{R}^{q}$, $q:=d(d+1)/2$, $t\in \mathbb{Z}$, instead (where the vech operator stacks the columns of a given square matrix from the principal diagonal downwards).}

{For our experiment, we use real stock market data. More explicitly, we collect the daily closing adjusted prices of the top $d=5, 10, 15$ companies-constituents of the S\&P500 stock index ordered alphabetically over last five years spanning from April 9, 2019 until April 9, 2024\footnote{The ordered list of the used tickers consists of MMM, AOS, ABT, ABBV, ACN, AIY.DE, ADBE, AAP, AMD, AFL, A, APD, AKAM, ALK, ALB. For a given choice of $d$, the first $d$ tickers are used from this list.}. Next, we compute the log-returns for these assets and use a quasi-maximum likelihood estimator to obtain the parameters of the model  \eqref{ret}-\eqref{H} (see~\cite{McAleer2008, Pedersen2014} for the proof of consistency and asymptotic normality of the diagonal BEKK QMLE estimator). Once the estimated parameters are available, we generate a sample of the input innovations corresponding to $3760$ observations, out of which the first $80\%$-long part of the sample {($3008$ observations)} is reserved for the training, while the last $20\%$  {($752$ observations)} are kept for testing purposes. Once given a sample of the $d$-dimensional input innovations, the task is to learn the corresponding $q$-dimensional outputs. }

 {We compare the performance in this task of the Volterra kernel with some of the existing and widely used static and sequential kernel benchmarks based on three chosen metrics. Along with the Volterra kernel, we also consider the static linear, Gaussian (RBF), polynomial, and sigmoid kernels together with three kernels for sequential data, namely, the $N$-level truncated signature kernel {Sig($N$)}~\cite{KO:19}, the global alignment kernel (GAK)~\cite{Cuturi2011}, and the signature PDE (Sig-PDE) kernel recently introduced in~\cite{salvi2021signature}.} {We refer the reader to Appendix~\ref{Appendix 4} for the details of the Volterra kernel implementation for the I/O learning task.}
{For all the considered kernels, we solve a kernel regularized regression as in \eqref{eq:representer mse}. We notice that each of the kernels has its own tunable hyperparameters. In our experiment, to ensure a fair comparison, we allocated a time budget of approximately $30$ minutes for hyperparameter tuning and subsequent model fitting of the best-selected model. The hyperparameter selection is conducted via the expanding window $5$-fold cross-validation (CV) using the $R^2$-score to measure the model goodness on the left-out folds. The time budget is used to determine the number of tuples $N_{\rm draw}$ in the hyperparameter space that can be afforded by CV in terms of the required time resource. Finally, to account for the fact that practitioners have no prior knowledge about the distribution of hyperparameters, they are randomly log-uniformly sampled to form $N_{\rm draw}$ combinations.}

{We report in Table~\ref{tab:1} three performance metrics for all the considered kernels and three different input dimensions. More precisely, for a given length of the testing sample $T_{\rm test}$ and for the samples of the true $\mathbf{h}'=\{\mathbf{h}'_t\}_{t=1}^{T_{\rm test}}$ and the model outputs $\widehat{\mathbf{h}}=\{\widehat{\mathbf{h}}_t\}_{t=1}^{T_{\rm test}}$, respectively, we evaluate the median absolute  error (MdAE)}
\begin{equation*}
	{\rm MdAE} (\mathbf{h}', \widehat{\mathbf{h}}) = \dfrac{1}{q}\sum_{i=1}^q{\rm med}\left(\left\{\left|{{h}'_{t,i}-\widehat{{h}}_{t,i}}\right|\right\}_{t=1}^{T_{\rm test}}\right),
\end{equation*}
{the normalized mean squared error (NMSE)}
\begin{equation*}
	{\rm NMSE} (\mathbf{h}', \widehat{\mathbf{h}}) = 
	\frac{1}
	{q T_{\rm test}}
	\sum_{i=1}^q\sum_{t=1}^{T_{\rm test}}{({h}'_{t,i}-\widehat{{h}}_{t,i})^2}/{ \sigma^2_{{h}', i}},
\end{equation*}
{with $\sigma^2_{{h}', i} = 
\frac{1}{ T_{\rm test}}
	\sum_{t=1}^{T_{\rm test}}
	({h}'_{t,i} - \mu_{{h}', i})^2$, $\mu_{{h}', i}=\frac{1}
	{ T_{\rm test}}\sum_{t=1}^{T_{\rm test}}{h}'_{t,i}$, and the Wasserstein-1 distance ($W_1$) between empirical cumulative distribution functions  $P_{h'}$ of $\mathbf{h}'$ and $Q_{\widehat{h}}$ of $\widehat{\mathbf{h}}$, given by}
\begin{equation*}
	W_1 (P_{h'}, Q_{\widehat{h}}) = 
	\sum_{t=1}^{T_{\rm test}}\|\mathbf{h}'_{(t)}-\widehat{\mathbf{h}}_{(t)}\|,
\end{equation*}
{where, for $t=1,\ldots, T_{\rm test}$,  $\mathbf{h}'_{(t)}$ and $\widehat{\mathbf{h}}_{(t)}$ are the $t$-th order statistics of the samples $\mathbf{h}'$ and $\widehat{\mathbf{h}}$, respectively.}

\begin{table}[ht]

\centering
\caption{{Median absolute error (MdAE)  scaled by 100, normalized mean squared error (NMSE), and Wasserstein-1 distance ($W_1$) scaled by 100, evaluated on the testing sample. {The models included in the NMSE model confidence set at $75\%$ confidence level (corresponding to the significance level $\alpha=0.25$)  are marked with $\ast\ast$.}}}
\label{tab:1}\resizebox{9cm}{!} 
{ 
	\begin{tabular}{ccccccccl}
\hline {\scriptsize Kernel} & {\scriptsize Linear}& {\scriptsize RBF} & {\scriptsize Poly} & {\scriptsize Sigmoid} & {\scriptsize GAK} & {\scriptsize {Sig($N$)}} & {\scriptsize Sig-PDE} & {\scriptsize Volterra}\\
\hline \multicolumn{8}{c}{$d=5$ and $q=15$}\\
\hline {\scriptsize MdAE}& {\scriptsize $1.971$}
 & {\scriptsize$1.819$}& {\scriptsize$1.828$}& {\scriptsize$1.822$} & {\scriptsize$2.298$}&{\scriptsize$1.961$}&{\scriptsize$1.816$} & {\scriptsize$\mathbf{1.033}$} \\
\hline {\scriptsize NMSE} & {\scriptsize $1.022$} & {\scriptsize$0.844$} & {\scriptsize$0.846$} & {\scriptsize$0.846$} & {\scriptsize$1.519$}&{\scriptsize$1.007$}&{\scriptsize$0.843$}&{\scriptsize$\mathbf{0.296}^{\ast\ast}$}\\
\hline {\scriptsize $W_1$} & {\scriptsize $2.356 $} & {\scriptsize$1.538 $} & {\scriptsize$ 1.564$} & {\scriptsize$ 1.523$} & {\scriptsize$2.235$} &{\scriptsize$2.266$}&{\scriptsize$1.559$}&{\scriptsize$\mathbf{0.455}$}\\
\hline \multicolumn{8}{c}{$d=10$ and $q=55$}\\
\hline {\scriptsize MdAE}& {\scriptsize $2.205$} & {\scriptsize $2.030$} & {\scriptsize $2.082$} & {\scriptsize $2.036$} & {\scriptsize $13.818$} & {\scriptsize $2.195$} & {\scriptsize $2.035$} & {\scriptsize $\mathbf{1.565}$}\\
\hline {\scriptsize NMSE} &{\scriptsize $1.079$}  & {\scriptsize $0.928$} &  {\scriptsize $0.941$}&  {\scriptsize $0.937$}& {\scriptsize $24.457$} & {\scriptsize $1.079$} & {\scriptsize $0.931$} & {\scriptsize $\mathbf{0.621}^{\ast\ast}$}\\
\hline {\scriptsize $W_1$} & {\scriptsize $2.659 $} & {\scriptsize$1.825 $} & {\scriptsize$ 2.034$} & {\scriptsize$ 1.712$} &{\scriptsize$13.696$}&{\scriptsize$2.628$} &{\scriptsize$1.796$}&{\scriptsize$\mathbf{1.025}$}\\
\hline \multicolumn{8}{c}{$d=15$ and $q=120$}\\
\hline {\scriptsize MdAE}& {\scriptsize $ 1.809$} & {\scriptsize $1.646$}&{\scriptsize $1.665$} & {\scriptsize $ 1.661$} & {\scriptsize $3.177 $} & {\scriptsize $ 1.806 $} & {\scriptsize $1.653$}& {\scriptsize $\mathbf{1.393}$}\\
\hline {\scriptsize NMSE} & {\scriptsize$1.041$} & {\scriptsize $0.866$ }&{\scriptsize $0.867$}  & {\scriptsize $0.881$} &{\scriptsize $2.245$} &{\scriptsize $1.041$}&{\scriptsize $0.872$}&{\scriptsize $\mathbf{0.623}^{\ast\ast}$}\\
\hline {\scriptsize $W_1$} & {\scriptsize $ 2.220 $} & {\scriptsize$ 1.369 $} & {\scriptsize$ 1.437 $} & {\scriptsize$ 1.199$} & {\scriptsize$ 3.248$}&{\scriptsize$2.189$}&{\scriptsize$1.355$}&{\scriptsize$\mathbf{ 0.932}$}\\
\hline
\end{tabular}
}
\end{table}
{Table~\ref{tab:1} demonstrates that the Volterra kernel offers superior performance compared to all the benchmarks with respect to all the performance metrics. Interestingly, the Wasserstein-1 metric shows the ability of our proposed kernel to capture the dynamics of the process better than its analogs. {In order to verify that the superior performance of our proposed Volterra with respect to other competing kernels is statistically significant, we apply the Model Confidence Set (MCS) test which is derived in~\cite{Hansen2011} for the case of mean squared losses. We establish that the Volterra kernel is the only model included in the set of models of equal predictive performance at $75\%$ confidence level. This level choice is informed by standard choices of the significance level $\alpha$ used in the literature. We intentionally take $\alpha=0.25$, which allows selecting a much smaller set of models with equal predictive power. Still, the Volterra kernel-based prediction happens to be the only model selected in the model set at this significance level.}}

{To illustrate the computational efforts associated with each of the studied kernels, we provide in Table~\ref{tab:2} the time in seconds required for the CV procedure for $N_{\rm draw}=100$ hyperparameter combinations for the learning task in dimension $d=10$ when run on a $120$ CPU cores server. {Additionally, we provide the time complexity of the kernel evaluation for each of the considered kernels.}} 
\begin{table}[ht]
\centering
\caption{Number of parameters of each kernel technique  (hyperparameters and regularization strength $\lambda_{\rm reg}$), {  the computational time required for CV, and the time complexity for computing one entry of the corresponding Gram matrix.} }
\label{tab:2}\resizebox{9cm}{!} 
{ 
	\begin{tabular}{ccccccccc}
\hline {\scriptsize Kernel} & {\scriptsize Linear}& {\scriptsize RBF} & {\scriptsize Poly($p$)} & {\scriptsize Sigmoid} & {\scriptsize GAK} & {\scriptsize {Sig($N$)}} & {\scriptsize Sig-PDE} & {\scriptsize Volterra}\\
\hline {\scriptsize Parameters} & {\scriptsize $1$}& {\scriptsize $2$} & {\scriptsize $4$} & {\scriptsize $3$} & {\scriptsize $2$} & {\scriptsize $4$} & {\scriptsize $3$} & {\scriptsize $3$}\\
\hline {\scriptsize Time (sec)}& {\scriptsize $ 2.47$}
 & {\scriptsize$ 3.28$}& {\scriptsize$ 2.89 $}& {\scriptsize$ 64.31$} & {\scriptsize$ 38817.75$}& {\scriptsize$ 23.92$}&{\scriptsize$2534.13 $} & {\scriptsize$ 21.36 $} \\
 \hline {{\scriptsize Complexity}}& {{\scriptsize$O(n d)$}}
 & {{\scriptsize$O(n d)$}} & {{\scriptsize$O(n d)$}}& {{\scriptsize$O(n d)$}} & {{\scriptsize$O(n^2 d)$}}& {{\scriptsize$O(n^2 N d)$}}&{{\scriptsize$O(n^2 d)$}} & {{\scriptsize$O(n d)$}} \\
\hline 
\end{tabular}}
\end{table}

{We emphasize that the numerical effort associated with the learning task using the Volterra kernel is much lower than the one invested in the use of other sequential kernels. For example, for Sig-PDE, a numerical solution for a Goursat-type PDE needs to be computed, while the Volterra kernel matrix is obtained via recursive expression \eqref{recursion gramian}. To our knowledge, such a methodology is not available for any of the other sequential kernels. It is important to mention that some substantial effort needs to be invested in choosing the support of the hyperparameters appropriately and in exploring the hyperparameter space more thoroughly whenever a particular learning problem is faced in practice.}


The code and the data necessary to reproduce these numerical results are publicly available at \url{https://github.com/Learning-of-Dynamic-Processes/reservoirkernel}.

{
Further comparisons of the Volterra kernel to other state-of-the-art time series kernels are carried out in~\cite{cass2024variance}. In~\cite{cass2024variance} the performance of these kernels is compared on the task of semi-supervised multivariate time series anomaly detection with data from the UEA multivariate time series repository~\cite{2018MultivateTimeSeriesRepository, 2021MultivariateBakeoff}. The Volterra  kernel achieves excellent performance in this task, with best precision-recall AUC (PR AUC) on many datasets. We refer to Section~4 in~\cite{cass2024variance} for further details. 
}
{Additionally, the performance of our proposed Volterra kernel is assessed for Lorenz and Mackey-Glass chaotic dynamical systems and compared to the polynomial kernel in~\cite{RC29}.}

\section{Conclusions}

{In this paper, we have constructed a universal kernel whose sections approximate any causal and time-invariant filter in the fading memory category with input and output {sequences composed out of elements in a finite-dimensional Euclidean space}. This kernel is built using as feature map the reservoir functional associated with a state-space system that naturally arises when writing the Volterra series expansion of any analytic fading memory filter as a linear image of the reservoir functional of a strongly universal infinite-dimensional state-space system.  The term strongly universal means that the state-space system employed in this realization does not depend on the FMP filter that is being realized. We refer to this infinite-dimensional system as the Volterra reservoir. The kernel map constructed using its reservoir functional is called the Volterra reservoir kernel.}

{We have shown that, even though the state-space representation and the corresponding reservoir feature map are defined on an infinite-dimensional tensor algebra space, the kernel map is characterized by explicit recursions that are readily computable for specific data sets when employed in estimation problems using the representer theorem. }

{Finally, }{we have showcased the performance of the Volterra reservoir kernel in a popular econometrics application in relation to the learning of multidimensional volatilities of financial stock returns.}

{In future research we intend to }{study in detail the relation between the objects that have been introduced here and what we believe is the continuous-time counterpart of our results, namely, the signature process in rough path theory, which plays a role very similar to that of the Volterra reservoir and for which a kernel representation can also be constructed and characterized as the solution of a Goursat partial differential equation}, {see~\cite{salvi2021signature}.} We find it also interesting to explore the connections with some recent work regarding more general choices of input and output spaces (see, for example,~\cite{galimberti2022designing} and~\cite{cuchiero2023global}, where sequence spaces of elements in Fr\'echet spaces and in infinite-dimensional weighted spaces are considered, respectively).


\section*{Acknowledgment}
We thank Christa Cuchiero, Christopher Salvi, and Josef Teichmann for interesting exchanges that inspired some of the results in this paper. JPO acknowledges partial financial support from the Swiss National Science Foundation (grant number 200021\_175801/1) and from the School of Physical and Mathematical Sciences of the Nanyang Technological University.

\ifCLASSOPTIONcaptionsoff
  \newpage
\fi



\bibliographystyle{myIEEEtran}
\bibliography{GOLibrary}

\appendix

\subsection{Kernel universality}
\label{Kernel universality}

Consider the RKHS $\mathbb{H}  $ associated with a continuous kernel map $K:{\cal Z}^{\Bbb Z_-} \times {\cal Z}^{\Bbb Z_-} \longrightarrow  \mathbb{R}$ as defined in \eqref{definition kernel map}. For any compact subset ${\mathcal K} \subset {\cal Z}^{{\Bbb Z}_-} $ we define, as in~\cite{micchelli2006universal}, the {\bfi  space of kernel sections} $K \left({\mathcal K}\right) $ consisting of all the functions in $C ^0({\mathcal K})$ which are uniform limits of functions of the form  
\begin{equation}
\label{def for kernel sections}
\sum_{i=1}^n a _iK _{{\bf z} _i},\, \mbox{with $a _i \in \mathbb{R}$,  $ {\bf z}_i \in {\mathcal K} $, $ n \in \mathbb{N} $.}
\end{equation} 
The kernel $K$ is called {\bfi  universal} (see~\cite{steinwart2001influence, micchelli2006universal}) when 
\begin{equation}
\label{universal kernel general}
K \left({\mathcal K}\right) =C ^0 \left({\mathcal K}\right),\, \mbox{for any compact subset ${\mathcal K} \subset {\cal Z}^{{\Bbb Z}_-} $.}
\end{equation}
It is easy to verify that $K \left({\mathcal K}\right) $ contains the completion of the space defined by \eqref{def for kernel sections}. Indeed, the reproducing property, the continuity of $K$, and the compactness of ${\mathcal K} $ imply that the uniform closure that defines $K \left({\mathcal K}\right) $ contains the completion of the vector space formed by the elements in \eqref{def for kernel sections}.
We emphasize that the universality results for reservoir families in, for instance,~\cite{RC6, RC7, RC8, RC20}, or~\cite{RC12} do not guarantee the universality of the associated reservoir kernels for each of the members of those families in the sense that we just stated.

\subsection{Kernelizing the estimation of linear  readouts}
\label{Kernelization Appendix}
A common estimation problem that appears when using systems of the form \eqref{rc state eq}-\eqref{rc readout eq} in practice and where the readout is linear (that is, in \eqref{rc readout eq} one has $h( \mathbf{x})= \left\langle {\bf W},  \mathbf{x}\right\rangle_{{\cal X}}$, for some ${\bf W} \in {\cal X} $), is finding the readout vector ${\bf W} \in {\cal X} $ that minimizes the empirical risk associated to a prescribed loss function $L: {\cal Y}\times {\cal Y} \rightarrow\mathbb{R}$ with respect to a finite sample of input/output observations. This is typically how one proceeds in reservoir computing, where the state equation is fixed and only a linear observation equation is subjected to training. In that particular case and if a quadratic loss is used, the estimation problem reduces itself to a (eventually regularized) regression problem with as many covariates as the dimension of the state space ${\cal X} $, which is in most cases very large. It is in this context that for quadratic or more general losses, the possibility of reducing the dimensionality of the estimation problem to the dimension of ${{\cal X}_R} $ using the RKHS technology that we just introduced may prove computationally advantageous. 

The next proposition is a slight generalization of the one that can be found in~\cite{RC16} formulated for finite-dimensional Hilbert state spaces ${\cal X} $ and shows two main consequences of the RKHS formulation of the estimation problem. First, even though the optimization problem that provides the optimal readout is formulated initially in the space $\mathcal{X} $, {\it it can be reduced to the dimensionally smaller} ${{\cal X}_R} $. Second, the Representer Theorem~\cite[page 117]{Mohri:learning:2012} shows that {\it the optimal readout is in the ``span of the data"}; this is the well-known ``kernelization trick" that in our case is computationally relevant in the presence of state spaces of dimension larger than the sample size or, as we shall see in the next section, infinite-dimensional state spaces. An important observation is that this second result {\it automatically yields a solution in the span ${{\cal X}_R} $ of the reachable set ${\cal X}_R $ without actually having to compute it}. We emphasize that the characterization of ${\cal X}_R $ is, in general, a difficult problem that limits the advantages of using ${\cal X}_R $ instead of ${\cal X}  $  as a dimensionality reduction technique.

We now introduce the different elements that are necessary for the statement of the proposition. First, we will assume that the state system $F: {\cal X} \times {\cal Z} \rightarrow\mathcal{X} $ is fixed and satisfies the ESP, and we are provided with a finite sample 
$\{( {\bf Z}_{-i}, Y_{-i})\}_{i\in \left\{ 0, \dots, n-1\right\}}$ of size $n$ of input/output observations. All along this section, the output observations $Y_{-i}$  are one-dimensional. For each time step $i\in \{ 0, \dots, n-1\}$ we define the truncated input samples as
\begin{equation} 
\label{finite sample}
{\bf Z}_{-i}^{-n+1} := (\ldots,{\bf 0},{\bf 0},{\bf Z}_{-n+1},\ldots,{\bf Z}_{-(i-1)},{\bf Z}_{-i}),
\end{equation}
that we use to define the {\bfi  training error} or the {\bfi  empirical risk} $\widehat{R}_n\left(H^F_{{\bf W}}\right)$ associated to the loss $L: {\cal Y}\times {\cal Y} \rightarrow\mathbb{R}$ for the system $H^F_{{\bf W}} (\cdot )= \langle{\bf W}, H^F(\cdot )\rangle_{{\cal X}}$ with readout  ${\bf W} \in {\cal X} $ as
\begin{equation}
\label{empirical risk}
\widehat{R}_n\left(H^F_{{\bf W}}\right) = \frac{1}{n} \sum_{i=1}^{n} L\left( \left\langle{\bf W}, H^F({\bf Z}_{-(i-1)}^{-n+1})\right\rangle_{{\cal X}},Y_{-(i-1)}\right).
\end{equation}
{
\begin{proposition}[Implicit reduction and kernelization]
\label{Implicit reduction and kernelization}
Let $F: {\cal X} \times {\cal Z} \rightarrow\mathcal{X} $ be a state system that satisfies the ESP and let $L: {\cal Y} \times {\cal Y}\rightarrow \mathbb{R}$ be a loss function with respect to the one-dimensional output space ${\cal Y} $. Let $\{( {\bf Z}_{-i}, Y_{-i})\}_{i\in \left\{ 0, \dots, n-1\right\}}$ be a sample of size $n$ of input/output observations. Let $\Omega: (0,\infty) \rightarrow \mathbb{R} $ be a strictly increasing function. Then, the regularized empirical risk minimization problem associated to $\widehat{R}_n $ admits the following potentially dimension-reducing reformulations:
\begin{itemize}[leftmargin=*]
\item {\bf Implicit reduction:} Let  $\mathbb{H}  $ be the RKHS introduced in \eqref{definition H for F} and characterized in \eqref{first identity H}. Then,
\begin{align} \label{optimizationProblem}
\min_{{\bf W}  \in {\cal X}}  &  \left\{\widehat{R}_n\left(H^F_{{\bf W}}\right)+ \Omega\left(\left\|{\bf W}\right\|_{{\cal X}}^2\right)\right\}\\
&=
\min_{{\bf W}  \in {{\cal X}_R}} \left\{\widehat{R}_n\left(H^F_{{\bf W}}\right)+ \Omega(\left\|{\bf W}\right\|_{{\cal X}}^2)\right\} \nonumber \\
	&=
\min_{H^F_{{\bf W}} \in \mathbb{H}} \left\{\widehat{R}_n\left(H^F_{{\bf W}}\right)+ \Omega(\left\|H_F^{{\bf W}}\right\|_{\mathbb{H} }^2)\right\}.\nonumber\end{align} 
Due to \eqref{first identity H}, these minima coincide with 
\begin{equation}
\label{min in rkhs}
\min_{f \in \mathbb{H}} \left\{\widehat{R}_n\left(f\right)+ \Omega(\left\|f\right\|_{\mathbb{H} }^2)\right\}.
\end{equation}
\item {\bf Kernelization:} In that case, the minimum in \eqref{min in rkhs} is realized by an element  
$H^F_{\widehat{{\bf W}}}= \left\langle \widehat{{\bf W}}, H ^F(\cdot )\right\rangle_{{\cal X}} \in \mathbb{H} $, with 
\begin{equation*}
\widehat{{\bf W}} =\sum_{i=1}^{n} \alpha _{i}H ^F({{\bf Z}_{-(i-1)}^{-n+1}}) \in {\cal X}
\end{equation*}
for some $\alpha_1, \alpha_2, \ldots, \alpha_{n} \in \mathbb{R}$ or, equivalently,
\begin{equation*}
H^F_{\widehat{{\bf W}}}=\underset{f\in \mathbb{H}}{\mbox{{\rm arg min}}} \left\{\widehat{R}_n\left(f\right)+ \Omega(\left\|f\right\|_{\mathbb{H} }^2)\right\}
=\sum_{i=1}^{n} \alpha _iK_{{\bf Z}_{-(i-1)}^{-n+1}}.
\end{equation*}
This implies that the optimization problem in \eqref{optimizationProblem}  in the space ${\cal X} $  can be reformulated in terms of the Gram matrix ${\mathcal K} \in \mathbb{M} _n $ of the sample  defined by
\begin{equation}
\label{gram matrix definition}
{\mathcal K} _{i,j}=K\left({\bf Z}_{-(n-i)}^{-n+1}, {\bf Z}_{-(n-j)}^{-n+1}\right), \quad \mbox{$i,j \in \left\{1, \ldots, n\right\}$}
\end{equation}
such that the optimization takes place in the space of vectors  $\boldsymbol{\alpha} \in \mathbb{R}^n$ with $n$ the sample size. More specifically, the optimization problem  \eqref{min in rkhs} is equivalent to
\begin{equation}
\label{optimization in terms of gramian}
\min_{\boldsymbol{\alpha} \in \mathbb{R} ^n} \bigg\{\frac{1}{n}\sum _{i=1}^n L \left(\left({\mathcal K}\boldsymbol{\alpha}\right)_i, Y_{-{(n-i)}}\right)+ \Omega(\boldsymbol{\alpha} ^{\top} {\mathcal K} \boldsymbol{\alpha})\bigg\}.
\end{equation}
\end{itemize}
\end{proposition}}
{
\begin{remark}[Kernelization for the empirical risk associated to the squared loss]{
	Consider the case in which the empirical risk in \eqref{empirical risk} is defined using the squared loss.
Let the regularization function $\Omega: (0,\infty) \rightarrow \mathbb{R} $ be defined as $\Omega(u) = \lambda u$, for some regularization strength $\lambda_{\rm reg}>0$ and let $\boldsymbol{Y}_n = (Y_{-(n-1)}, Y_{-(n-2)}, \ldots, Y_0)^\top$. Then, the optimization problem \eqref{optimization in terms of gramian} in Proposition~\ref{Implicit reduction and kernelization}  can be rewritten as\begin{equation*}\min_{\boldsymbol{\alpha} \in \mathbb{R} ^n} \left\{\| \mathcal{K}\boldsymbol{\alpha} - \boldsymbol{Y}_n \|_2^2+ \lambda_{\rm reg} \boldsymbol{\alpha} ^{\top} {\mathcal K} \boldsymbol{\alpha} \right\},
 \end{equation*}}
which admits the unique {\it closed-form} solution given by 
\begin{equation}
\label{eq:representer mse}
\boldsymbol{\alpha}^\ast = \left(\mathcal{K}^{\top} \mathcal{K}+\lambda_{\rm reg} \mathcal{K}\right)^{-1} \mathcal{K} \boldsymbol{Y}_n.
 \end{equation}
\end{remark}}

The implicit reduction in Proposition~\ref{Implicit reduction and kernelization} constitutes an advantage whenever the dimension of  ${ {\cal X} _R}$ of the reachable set of the system is smaller than that of ${\cal X}$. Regarding kernelization, two requirements should be satisfied for this method to be advisable, namely, the sample size should be smaller than the dimension of ${\cal X}$ and, additionally, it should be possible to easily compute the Gram matrix \eqref{gram matrix definition}. This last requirement may be problematic for nonlinear state systems since, in general, an explicit expression of the functional $H ^F $ is not available. This  makes it obviously difficult to evaluate the associated kernel map \eqref{definition kernel map}. This issue has been addressed in Section~\ref{The reservoir kernel of a Volterra series} for the Volterra kernel via sequentialization.

\subsection{Proof of Theorem~\ref{The Volterra reservoir}}
\label{Appendix 1}
\noindent Part \textbf{(i)} 
%
is a consequence of the fact that,  using the norm in $T_2(\mathbb{R}^d) $ induced by the inner product introduced in \eqref{dif inner prod prod}, the Volterra reservoir map  $F^{{\rm Volt}}_\lambda(\mathbf{x}, {\bf z})=\lambda \mathbf{x}   \otimes \widetilde{{\bf z}}  + 1 $ is a contraction on the state variable with constant  $\lambda/ \sqrt{1- \tau ^2 M ^2} <1$, due to the hypothesis on $\lambda $ in the statement, that is,
\begin{equation*}
	\left\|F^{{\rm Volt}}_\lambda(\mathbf{x}, {\bf z})-F^{{\rm Volt}}_\lambda({\bf y}, {\bf z})\right\|\leq \frac{\lambda }{\sqrt{1- \tau ^2 M ^2} } \left\|\mathbf{x}- {\bf y}\right\|
\end{equation*}
for all $\left\|{\bf z}\right\|\leq M $.
This fact, together with {the same arguments as in~\cite[Proof of Proposition~1]{RC10},~\cite[Theorem 3.1]{RC7}}, proves the existence of the Volterra filter $U ^{{\rm Volt}}  _\lambda $ and that it has the FMP, see also~\cite[Remark~2]{RC10}. Part {\bf (ii)} is a straightforward corollary of Theorem~\ref{volterra representation} and of the expression \eqref{solution uvolt} of the filter $U ^{{\rm Volt}}  _\lambda$ which contains all the monomials of all polynomial orders on all the variables ${\bf z} _t $, $t \in \mathbb{Z}_{-} $. The linear map ${\bf W}$ is constructed by matching the coefficients $g _j(m _1, \ldots, m _j) $ of the  Volterra series representation of $U$ with the terms of the filter $U ^{{\rm Volt}}  _\lambda $ in \eqref{solution uvolt}. \quad $\blacksquare$

\subsection{Proof of Proposition~\ref{The Volterra reservoir kernel proposition}}
\label{Appendix 2}
\noindent Using the definition of the kernel map \eqref{definition kernel map}, the expression \eqref{Volterra reservoir eq} that defines the Volterra reservoir, and its time-invariance, we obtain that
\begin{multline*}
	K^{{\rm Volt}}({\bf z}, {\bf z}')= \left\langle U ^{{\rm Volt}}  _\lambda({\bf z})_0, U ^{{\rm Volt}}  _\lambda({\bf z}')_0 \right\rangle\\
	= \left\langle 1+\lambda U ^{{\rm Volt}}  _\lambda({\bf z})_{-1} \otimes \widetilde{ {\bf z}} _0, 1+\lambda U ^{{\rm Volt}}  _\lambda({\bf z}')_{-1} \otimes \widetilde{ {\bf z}'} _0 \right\rangle\\
	=1 +\lambda ^2\left\langle  U ^{{\rm Volt}}  _\lambda(T _1({\bf z}))_{0},  U ^{{\rm Volt}}  _\lambda(T _1({\bf z}'))  _0 \right\rangle  \left\langle \widetilde{ p _0({\bf z})} ,\widetilde{ p _0({\bf z}')}\right\rangle\\
	=1+ \frac{\lambda ^2  K^{{\rm Volt}}(T _1({\bf z}), T _1( {\bf z}'))}{1- \tau ^2\langle p _0({\bf z}), p _0({\bf z}') \rangle},
\end{multline*}
which proves \eqref{recursion for kvolt}. In the last equality we used \eqref{norm of tilde} together with the fact that $ {\bf z}, {\bf z}' \in K _M  $ and the hypothesis $\tau^2 M ^2<1 $. The expression \eqref{recursion for kvolt sum} is the unique solution of \eqref{recursion for kvolt} because for any ${\bf z}, {\bf z}' \in K _M $, $(1- \tau ^2\langle p _0({\bf z}), p _0({\bf z}') \rangle)^{-1 }\leq (1- \tau ^2M ^2)^{-1}$ and since, by hypothesis, $0< \lambda< \sqrt{1- \tau ^2 M ^2} $, the recursion \eqref{recursion for kvolt} is necessarily contractive. On the other hand, $K^{{\rm Volt}}$ is bounded by $L^2$ (with $L$ defined in Theorem~\ref{The Volterra reservoir}~{\bf (i)}) and hence it has a unique solution. \quad $\blacksquare$

\subsection{Proof of Theorem~\ref{thm:universality}}
\label{Appendix 3}

\noindent Notice first of all that, by definition, the space of kernel sections $K ^{{\rm Volt}} (K _M) $ consists of all the functions which are uniform limits of functions of the form  $\sum_{i=1}^n a _iK^{{\rm Volt}}  _{{\bf z} _i}$, with $a _i \in \mathbb{R}$,  $ {\bf z}_i \in K _M $, $ n \in \mathbb{N} $. Each of the functions $K^{{\rm Volt}}  _{{\bf z} _i}(\cdot )= \left\langle H^{{\rm Volt}} _\lambda ({\bf z} _i), H^{{\rm Volt}} _\lambda (\cdot )\right\rangle $ is continuous because by Theorem~\ref{The Volterra reservoir} the Volterra functional $ H^{{\rm Volt}} _\lambda $ has the fading memory property and it is hence continuous. Since uniform limits of continuous functions are continuous, we can conclude that $K ^{{\rm Volt}} (K _M)\subset C ^0(K _M)$.

Equality is guaranteed by Theorem 7 in~\cite{micchelli2006universal}, which proves that $K ^{{\rm Volt}} $ is universal if one of its sets of features is universal. We recall that a set of features of a kernel associated with a given feature map are its functional coefficients in an orthonormal basis decomposition. We now show that this property is indeed satisfied by the Volterra functional $H ^{{\rm Volt}}  _\lambda: K _M \rightarrow T _2(\mathbb{R}^d)$ given by {
\begin{equation*}
	\label{solution hvolt}
	H^{{\rm Volt}}  _\lambda({\bf z})=1+\sum_{j=0}^{\infty} \lambda^{j+1}\bigotimes_{k=0}^j\widetilde{{\bf z}} _{-k} ,
\end{equation*}}
when expressed in terms of the canonical orthonormal basis of $T _2(\mathbb{R}^d)$. We make this statement explicit only for one-dimensional inputs ($d=1 $) in order to shorten the presentation. The extension to the general case is straightforward. Define, for any fixed $j \in \{0,1,\ldots \}$ and $i _0, i _1, \ldots, i _j \in \{0,1,\ldots \}$, the maps $\phi_{j, i _0, i_1,  \ldots, i _j}:K _M \rightarrow \mathbb{R}  $ given by
\begin{equation*}
	\phi_{j, i _0, i_1,  \ldots, i _j}({\bf z})= \lambda^{j+1} \tau^{i_0+\cdots+i_j} z _0^{i _0}z _{-1}^{i _1} \cdots z _{-j}^{i _j}, 
\end{equation*}
and note that {
\begin{multline*}
	H^{{\rm Volt}}  _\lambda({\bf z})=1+\sum_{j=0}^{\infty} \sum_{i _0=0}^{\infty} \cdots \sum_{i _j=0}^{\infty}
	\phi_{j, i _0, i_1,  \ldots, i _j}({\bf z}) \bigotimes_{k=0}^j \mathbf{e}^{\otimes^{i _k}} ,
\end{multline*}}
where the symbol $\mathbf{e}^{\otimes^{i _k}}$ denotes the simple tensor in $T^{i _k}( \mathbb{R}) $ obtained as the tensor product of $ i _k $ copies of the vector $\mathbf{e}=1 \in \mathbb{R}$ and the bigger symbol $\Motimes $ stands for the tensor product in $T_2( \mathbb{R}) $. Note that 
\begin{align*}
	K^{{\rm Volt}}({\bf z},& {\bf z}')=\\&1+\sum_{j=0}^{\infty} \sum_{i _0=0}^{\infty} \cdots \sum_{i _j=0}^{\infty}
	\phi_{j, i _0, i_1,  \ldots, i _j}({\bf z}) \phi_{j, i _0, i_1,  \ldots, i _j}({\bf z'}) 
\end{align*}
and that the orthogonormality of the basis elements $\left\{\mathbf{e}^{\otimes^{i _j}} \Motimes \mathbf{e}^{\otimes^{i _{j-1}}}\Motimes \cdots \Motimes \mathbf{e}^{\otimes^{i _{0}}}\right\} _{j, i _0, \ldots, i _j}$ implies that the set $\left\{1\right\}\cup \left\{\phi_{j, i _0, i_1,  \ldots, i _j}\right\}_{j, i _0, \ldots, i _j} $ forms a set of features for $K^{{\rm Volt}} $. The linear span of this set contains the constant maps, it separates the points in $K _M $ (it contains the monomials of order one in each of the components of the input sequence), and forms a polynomial algebra. Indeed, for two arbitrary  sets of indices $\left\{j, i _0, i_1,  \ldots, i _j\right\} $  and $\left\{ j', i _0', i_1',  \ldots, i _j' \right\} $, with $j'\geq j  $,  and any ${\bf z} \in K _M $, we have
\begin{multline*}
	\phi_{j, i _0, i_1,  \ldots, i _j}({\bf z}) \cdot \phi_{j', i _0', i_1',  \ldots, i _j'}({\bf z})\\
	=
	\lambda^{j+1}
\phi_{j', i _0+i _0', i _1+i_1',  \ldots, i _j+i _j', i_{j +1}', \ldots, i_{j'}'}({\bf z}).
\end{multline*}
The Stone-Weierstrass Theorem guarantees that this set of features is hence universal. \quad $\blacksquare$

\subsection{Implementation of numerical illustration in Section~\ref{Numerical illustration}}
\label{Appendix 4}
{
Consider a given sample of inputs $\{\mathbf{z}_t\}_{t\in \{1,\ldots, T\}}$, $\mathbf{z}_t\in \mathbb{R}^d$, and the associated sample of outputs $\{\mathbf{h}_t\}_{t\in \{1,\ldots, T\}}$, $\mathbf{h}_t\in \mathbb{R}^q$, with $q:=d(d+1)/2$  (in our experiment $T=3008$). The inputs are normalized such that $M=1$. For some chosen $\tau>0$ take $0<\lambda<\sqrt{1-\tau^2}$ as required in Theorem~\ref{The Volterra reservoir}, and initialize ${\mathcal K}^{{\rm Volt}} _{0,0}={\mathcal K}^{{\rm Volt}} _{i,0}=1 / (1- \lambda ^2)$ for all $i\in \left\{1, \ldots, T\right\}$. We recall that the kernel hyperparameters need to be carefully chosen depending on the learning task. To that aim, data-driven approaches prove useful, and one would use for example cross-validation techniques for a disciplined hyperparameter selection. Subsequently, one uses the recursion in~\eqref{recursion gramian} to compute the  entries of the Gram matrix as 
\begin{equation*}
{\mathcal K}^{{\rm Volt}} _{i,j}=1 + \lambda ^2 \frac{{\mathcal K}^{{\rm Volt}} _{i-1,j-1}}{1- \tau ^2  {\bf z}_{i}^{\top}  {\bf z}_{j}},
\end{equation*}
for all $i=1,\ldots, T$ and $j \in  \left\{1, \ldots, i\right\}$, and completes the matrix using its symmetry. This results in the Gram matrix $\mathcal{K}^{{\rm Volt}}\in \mathbb{S}_T$. One can proceed further by  organizing the target outputs into a matrix  $Y := (\mathbf{h}_1| \mathbf{h}_{2}| \cdots| \mathbf{h}_T)^\top \in \mathbb{M}_{T,q}$ and choosing the squared loss to define the empirical risk in \eqref{empirical risk}. Further, one states the empirical risk minimization problem which, by the Representer Theorem admits, for some regularization strength $\lambda_{\rm reg}>0$ (also subject to cross-validation), a  closed-form solution as in \eqref{eq:representer mse}, that is
\begin{equation*}
A^\ast:=(\boldsymbol{\alpha}^\ast_{1}| \boldsymbol{\alpha}^\ast_{2}|\cdots|\boldsymbol{\alpha}^\ast_{q})= \left(\mathcal{K}^2 +\lambda_{\rm reg} \mathcal{K}\right)^{-1} \mathcal{K} {Y} \in \mathbb{M}_{T,q},  \end{equation*}
 with $\boldsymbol{\alpha}^\ast_{k}\in \mathbb{R}^T$, $k=1,\ldots, q$. We note that in practical scenarios, it is customary to use only the right lower block of the Gram matrix, ensuring that the impact of the initialization is reduced and, as a byproduct, lowering the dimension of the optimization problem. This approach would result in  $\boldsymbol{\alpha}^\ast_{k}\in \mathbb{R}^{\widetilde{T}}$, $k=1,\ldots, q$, with $\widetilde{T}\ll T$. Consider now a new testing input sample $\{\mathbf{z}'_t\}_{t=1}^{T_{\rm test}}$, $\mathbf{z}'_t\in \mathbb{R}^d$, and the associated testing targets $\{\mathbf{h}'_t\}_{t=1}^{T_{\rm test}}$, $\mathbf{h}'_t\in \mathbb{R}^q$, with $q:=d(d+1)/2$ (in our case of length $T_{\rm test} = 752$) to be given. One computes the Gram matrix of the Volterra kernel between the training and testing inputs as follows
 \begin{equation*}
({\mathcal K}^{{\rm Volt}} _{T_{\rm test}})_{i,j}=1 + \lambda ^2 \frac{({\mathcal K}^{{\rm Volt}} _{T_{\rm test}}) _{i-1,j-1}}{1- \tau ^2  {\bf z}_{i}^{'\top}  {\bf z}_{j}}, 
\end{equation*}
for $i=1,\ldots, T_{\rm test}$, $j=1,\ldots, T$, 
 using the same initialization as in the training phase, that is $({\mathcal K}^{{\rm Volt}} _{T_{\rm test}}) _{0,0}=({\mathcal K}^{{\rm Volt}} _{T_{\rm test}}) _{i,0}=({\mathcal K}^{{\rm Volt}} _{T_{\rm test}}) _{0,j}=1 / (1- \lambda ^2)$ for all $i\in \left\{1, \ldots, T_{\rm test}\right\}$ and for all $j\in \left\{1, \ldots, T\right\}$. One immediately obtains the predicted outputs computing
 \begin{equation*}
 	\widehat{Y}_{T_{\rm test}} := (\widehat{\mathbf{h}}_1| \widehat{\mathbf{h}}_{2}| \cdots| \widehat{\mathbf{h}}_{T_{\rm test}})^\top= {\mathcal K}^{{\rm Volt}} _{T_{\rm test}} A^\ast \in \mathbb{M}_{T_{\rm test},q}.
 \end{equation*}
Finally, one can assess the performance of the Volterra kernel by comparing the testing targets $\{\mathbf{h}'_t\}_{t=1}^{T_{\rm test}}$ with the predicted outputs $\{\widehat{\mathbf{h}}_t\}_{t=1}^{T_{\rm test}}$ using any preferred performance metrics (see Section~\ref{Numerical illustration} for the details).}


%








\end{document}